\documentclass[11pt,a4paper]{article}
\usepackage[hyperref]{emnlp2020}
\usepackage{times}
\usepackage{lipsum}
\usepackage{latexsym}
\usepackage[ruled,linesnumbered]{algorithm2e}
\usepackage{amsmath,amssymb,graphicx,wasysym}
\usepackage{multirow,booktabs,xcolor,colortbl}
\usepackage{tabularx}

\usepackage{microtype}

\providecommand\rightarrowRHD{\relbar\joinrel\mathrel\RHD}
\providecommand\rightarrowrhd{\relbar\joinrel\mathrel\rhd}
\providecommand\longrightarrowRHD{\relbar\joinrel\relbar\joinrel\mathrel\RHD}
\providecommand\longrightarrowrhd{\relbar\joinrel\relbar\joinrel\mathrel\rhd}

\makeatletter
\providecommand*\xrightarrowRHD[2][]{\ext@arrow 0055{\arrowfill@\relbar\relbar\longrightarrowRHD}{#1}{#2}}
\providecommand*\xrightarrowrhd[2][]{\ext@arrow 0055{\arrowfill@\relbar\relbar\longrightarrowrhd}{#1}{#2}}
\makeatother

\aclfinalcopy 
\def\aclpaperid{1402}
\def\ie{i.e.\ }

\newcommand\ra{$\rightarrow$}
\newcommand{\MC}[3]{\multicolumn{#1}{#2}{#3}}
\newcommand{\MR}[3]{\multirow{#1}{#2}{#3}}

\newcommand{\B}{\textbf}
\newcommand{\I}{\textit}
\newcommand{\T}{\texttt}
\newcommand{\U}{\underline}
\newcommand{\true}[1]{\textbf{#1}}
\newcommand{\false}[1]{\underline{{#1}}}

\newcommand{\ua}[1]{\textcolor{black}{\small$\Uparrow {#1}$}}
\newcommand{\da}[1]{\textcolor{red}{\small$\Downarrow {#1}$}}

\title{Simultaneous Machine Translation with Visual Context}

\author{Ozan Caglayan$^{1}${\normalfont,} Julia Ive$^{1}${\normalfont,} Veneta Haralampieva$^1${\normalfont,}  Pranava Madhyastha$^1$ \\[.2em] {\bf Lo\"ic Barrault$^2$ \and Lucia Specia$^{1,2,3}$}\\[.3em]
Imperial College London$^1$,\, University of Sheffield$^2$,\, ADAPT - Dublin City University$^3$\\
         \texttt{\small o.caglayan@ic.ac.uk, j.ive@ic.ac.uk, vlh19@ic.ac.uk, pranava@ic.ac.uk} \\ \texttt{\small l.barrault@sheffield.ac.uk, l.specia@ic.ac.uk}\\
}

\date{}

\begin{document}
\maketitle
\begin{abstract}
Simultaneous machine translation (SiMT) aims to translate a continuous input text stream into another language with the lowest latency and highest quality possible. The translation thus has to start with an incomplete source text, which is read progressively, creating the need for anticipation. In this paper, we seek to understand whether the addition of visual information can compensate for the missing source context. To this end, we analyse the impact of different multimodal approaches and visual features on state-of-the-art SiMT frameworks. Our results show that visual context is helpful and that visually-grounded models based on explicit object region information are much better than commonly used global features, reaching up to 3 BLEU points improvement under low latency scenarios. Our qualitative analysis illustrates cases where only the multimodal systems are able to translate correctly from English into gender-marked languages, as well as deal with differences in word order, such as adjective-noun placement between English and French.
\end{abstract}

\section{Introduction}
\label{sec:intro}
Simultaneous machine translation (SiMT) aims to reproduce human interpretation, where an interpreter translates spoken utterances as they are produced. The interpreter has to dynamically find the balance between how much context is needed to generate the translation reliably, and how long the listener has to wait for the translation. In contrast to \I{consecutive} machine translation where source sentences are available in their entirety before translation, the challenge in SiMT is thus the design of a  strategy to find a good trade-off between the quality of the translation and the latency incurred in producing it. Previous work has considered \I{rule-based strategies} that rely on waiting until some constraint is satisfied, which includes approaches based on syntactic constraints~\cite{bub1997verbmobil,ryu-etal-2006-simultaneous}, segment/chunk/alignment information~\cite{bangalore-etal-2012-real}
heuristic-based conditions during decoding~\cite{cho2016can} or deterministic policies with pre-determined latency constraints~\cite{ma-etal-2019-stacl}.
An alternative line of research focuses on \I{learning} the decision policy: \citet{gu-etal-2017-learning} and \citet{alinejad-etal-2018-prediction} frame SiMT as learning to generate \T{READ/WRITE} actions and employ reinforcement learning (RL) to formulate the problem as a policy agent interacting with its environment (\ie a pre-trained MT model). Recent work has also explored supervised learning of the policy, by using oracle action sequences predicted by a pre-trained MT using confidence-based heuristics~\cite{zheng-etal-2019-simpler} or external word aligners~\cite{arthur-et-al-learning} (details in $\S$\ref{sec:relwork}).

\begin{figure}[t!]%
\centering
\setlength{\belowcaptionskip}{-10pt}
\includegraphics[width=0.45\textwidth]{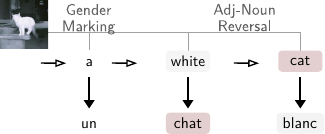}
\caption{An illustration of a 1-word latency system that makes use of visual grounding to resolve the gender of the article `un' and to predict the noun `chat' after reading its qualifier `white'. $\rightarrowrhd$ and $\rightarrowRHD$ denote \T{READ} and \T{WRITE}, respectively.}
\label{fig:intro}
\end{figure}
Thus far, all prior research has focused on unimodal interpretation\footnote{We note that \citet{imankulova2020multimodal} also attempted to explore multimodality in SiMT. However, their paper over-estimates the impact of visual cues, and in personal correspondence with the authors about the mismatch in the findings, they discovered critical bugs in their implementation.}. In this paper, we explore
SiMT for multimodal machine translation (MMT)~\cite{specia-etal-2016-shared}, where in addition to the source sentence, we have access to visual information in the form of an image. We believe that having access to a complementary context should help the models anticipate the missing context (Figure~\ref{fig:intro}) by grounding their decisions about `when' and `what' to translate.
To test our hypothesis, we explore heuristic-based decoding and fixed-latency wait-$k$ policy\footnote{During our initial experiments we also explored the RL-based SiMT policy~\cite{gu-etal-2017-learning} but could not find good hyper-parameter settings, especially settings which were stable across two language pairs. Therefore, we did not proceed with RL for multimodal SiMT.}, and investigate the effectiveness of different visual representations ($\S$\ref{sec:method}). Based on experiments on the Multi30k dataset~\cite{elliott-etal-2016-multi30k} ($\S$\ref{sec:setup}), we provide quantitative and qualitative analyses for English\ra German and English\ra French ($\S$\ref{sec:results}).

Our \B{contributions} highlight that: (i) visual context offers up to 3 BLEU points improvement for low-latency wait-$k$ policies, and consistently lowers the latency for \I{wait-if-diff}~\cite{cho2016can} decoding, (ii) explicit object region features are more expressive than commonly used global visual features, (iii) training wait-$k$ MMTs offers remarkably better grounding capabilities than decoding-only wait-$k$ for linguistic phenomena such as gender resolution and adjective-noun ordering, and (iv) with twice the runtime speed of decoder-based visual attention, the encoder-based grounding is promising for application scenarios.

\section{Related Work}
\label{sec:relwork}
\subsection{Multimodal Machine Translation (MMT)}
MMT aims to improve the quality of automatic translation using auxiliary sources of information~\cite{sulubacak2019multimodal}. The most typical framework explored in previous work makes use of the images when translating their descriptions between languages, with the hypothesis that visual grounding could provide contextual cues to resolve linguistic phenomena such as word-sense disambiguation or gender marking. 

Existing work often rely on the use of visual features extracted from state-of-the-art CNN models pre-trained on large-scale visual tasks. The methods can be grouped into two branches depending on the feature type used: (i) multimodal attention~\cite{calixto-etal-2016-dcu,caglayan-etal-2016-multimodality, libovicky-helcl-2017-attention,delbrouck2017modulating} which implements a soft attention~\cite{Bahdanau2014} over \I{spatial} feature maps, and (ii)
multimodal interaction between a \I{pooled} visual feature vector and linguistic representations~\cite{calixto-liu-2017-incorporating,caglayan-etal-2017-lium,elliott-kadar-2017-imagination,gronroos-etal-2018-memad}.

\subsection{Simultaneous Neural MT}
Simultaneous NMT was first explored by \citet{cho2016can} in a greedy decoding framework where heuristic waiting criteria are used to decide whether the model should read more source words or emit a target word. \citet{gu-etal-2017-learning} instead utilised a pre-trained NMT model in conjunction with a reinforcement learning agent whose goal is to learn a \T{READ/WRITE} policy by maximising quality and minimising latency. \citet{alinejad-etal-2018-prediction} further extended the latter approach by adding a \T{PREDICT} action whose purpose is to anticipate the next source word.

A common property of the above approaches is their reliance on \I{consecutive} NMT models pre-trained on full-sentences.
\citet{dalvi-etal-2018-incremental} pointed out a potential mismatch between the training and decoding regimens of such approaches and proposed fine-tuning the models using chunked data or prefix pairs.
\citet{ma-etal-2019-stacl} proposed an end-to-end, fixed-latency framework called `wait-$k$' which allows \I{prefix-to-prefix} training using a deterministic policy: the agent starts by reading a specified number of source tokens ($k$), followed by alternating 
\T{WRITE} and \T{READ} actions.
\citet{arivazhagan-etal-2019-monotonic} extended the wait-$k$ framework using an advanced attention mechanism and optimising a differential latency metric (DAL). Recently, \citet{arivazhagan-et-al-retranslation} explored a radically different approach which enriches full-sentence training with prefix pairs~\cite{niehues2018}
and allows \I{re-translation} of previously committed target tokens to increase the translation quality.

Another line of research focuses on learning adaptive policies in a supervised way by using oracle \T{READ/WRITE} actions generated with heuristic or alignment-based approaches.  \citet{zheng-etal-2019-simpler} extracted action sequences from a pre-trained NMT model with a confidence-based heuristic and used them to train a separate policy network while \citet{arthur-et-al-learning} explored jointly training the translation model and the policy with oracle sequences obtained from a word alignment model.

\section{Methods}
\label{sec:method}
In this section, we first describe the underlying NMT architectures and baseline simultaneous MT approaches, to then introduce the proposed multimodal extensions to SiMT.

\subsection{Baseline NMT}
\label{sec:method:baseline}
Our \textit{consecutive} baseline consists of a 2-layer GRU~\cite{cho-etal-2014-learning} encoder and a 2-layer Conditional GRU decoder~\cite{sennrich-etal-2017-nematus} with attention~\cite{Bahdanau2014}.
The encoder is \B{unidirectional} as source sentences would be progressively read\footnote{Although it is possible to encode the growing prefixes bidirectionally, it would incur a quadratic complexity.}.
Given a source sequence of embeddings $X\mathrm{=} \{x_1,\dots,x_S\}$ and a target sequence of embeddings $Y\mathrm{=}\{y_1,\dots,y_T\}$, the encoder first computes the sequence of hidden states $H\mathrm{=}\{h_1,\dots,h_S\}$. At a given timestep $t$ of decoding, the output layer estimates the probability of the next target word $y_t$ as follows: 
\begin{align}
d_t  &= \text{GRU}(y_{t-1}, d'_{t-1})\nonumber\\
c_t  &= \text{Attention}(H, \text{query}\leftarrow d_t)\label{eq:ctx}\\
d_t' &= \text{GRU}'(c_t, d_{t})\nonumber\\
o_t &= \tanh(\mathbf{W_c}c_t + \mathbf{W_d}d_t' + \mathbf{W_y}y_{t-1})\nonumber\\
l_t &= \mathbf{W_o}(\mathbf{W_b}o_t + b_b) + b_o\nonumber\\
P(y_t | X, Y_{<t}) &= \text{softmax}(l_t)\nonumber
\end{align}
For a single training sample, we then maximise the joint likelihood of source and target sentences:
\begin{align}
    \textsc{L}(X, Y) = \sum_{t=1}^T \log\left(P\left(y_t | X_{\leq g(t)}, Y_{< t}\right)\right)\label{eq:mle}
\end{align}
Following the formulation of \citet{ma-etal-2019-stacl}, $g(t)$ in equation~\ref{eq:mle} is a function which returns the number of source tokens encoded so far when predicting the target token $y_t$. In the case of consecutive NMT, since all source tokens are observed before the decoder runs, $g(t)$ is equal to the length of the source sentence \ie $g(t) = |X|$.

\subsection{Incorporating the visual modality}
We consider a setting where the visual context is static and is available in its entirety at encoding time. This is a realistic setting in many applications, for example, the simultaneous translation of news, where images (or video frames) are shown before the whole source stream is available. We consider the following ways of integrating visual information.

\paragraph{Object classification (OC) features} are \I{global} image information extracted from convolutional feature maps, which are believed to capture spatial cues. These \I{spatial} features are extracted from the final convolution layer of a ResNet-50 CNN~\cite{he2016resnet} trained on ImageNet~\cite{deng2009imagenet} for object classification. An image is represented by a feature tensor $V \in \mathbb{R}^{8\times 8\times 2048}$.

\paragraph{Object detection (OD) features} are explicit object information where \I{local} regions in an image detected as objects are encoded by \I{pooled} feature vectors. These features are generated by the ``bottom-up-top-down (BUTD)"~\cite{butd} extractor\footnote{https://hub.docker.com/r/airsplay/bottom-up-attention} which is a Faster R-CNN/ResNet-101 object detector (with 1600 object labels) trained on the Visual Genome dataset~\cite{visualgenome}. For a given image, the detector provides 36 object region proposals and extracts a pooled feature vector from each. An image is thus represented by a feature tensor $V \in \mathbb{R}^{36\times 2048}$. We hypothesise that explicit object information can result in better referential grounding by using conceptually meaningful units rather than global features.

\subsection{Multimodal architectures}
\paragraph{Decoder attention (DEC-OC/OD).}
A standard way of integrating visual modality into NMT is to apply a secondary attention at each decoding timestep~\cite{calixto-etal-2016-dcu,caglayan-etal-2016-multimodality}. We follow this approach to construct an MMT baseline. Specifically, equation~\ref{eq:ctx} is extended so that the decoder attends to both the source hidden states $H$ (eq.~\ref{eq:txtctx}) and the visual features $V$ (eq.~\ref{eq:visctx}), and they are added together to form the \textit{multimodal} context vector $c_t$ (eq.~\ref{eq:mmctx}):
\begin{align}
c_t^{\mathbf{T}} &= \text{Attention}^{\mathbf{T}}(H, \text{query=}d_t)\label{eq:txtctx}\\
c_t^{\mathbf{V}} &= \text{Attention}^{\mathbf{V}}(V, \text{query=}d_t)\label{eq:visctx}\\
c_t &= c_t^{\mathbf{T}} + c_t^{\mathbf{V}}\label{eq:mmctx}
\end{align}

\paragraph{Multimodal encoder (ENC-OD).} Instead of integrating the visual modality into the decoder, we propose to ground the source sentence representation within the encoder similar to \citet{delbrouck2017modulating}. We hypothesise that early visual integration could be more appropriate for SiMT to fill in the missing context. Our approach differs from \citet{delbrouck2017modulating} in the use of \I{scaled-dot} attention~\cite{vaswani-etal-2017-attention} and \B{object detection (OD)} features. The attention layer receives unidirectional hidden states $H$ (for source states that were encoded/read so far) as the \I{query} and the visual features $V$ as \I{keys} and \I{values}, \ie it computes a mixture $M$ of region features based on the cross-modal relevance. The final representation that will be used as input to the equation~\ref{eq:ctx} is defined as $\textsc{LayerNorm}(M + H)$ \cite{ba2016layer}.

Regardless of the multimodal approach taken, all visual features are first linearly projected into the dimension of textual representations $H$. To make modality representations compatible in terms of magnitude statistics, we apply layer normalisation~\cite{ba2016layer} on textual representations $H$ and the previously projected visual representations $V$.
A dropout~\cite{srivastava2014dropout} of $p=0.5$ follows the layer normalisation.

\subsection{Simultaneous MT approaches}
This section summarises the SiMT approaches explored in this work: (i) the \textit{heuristic-based} decoding approach \B{wait-if-diff}~\cite{cho2016can}, (ii) the wait-$k$ policy~\cite{ma-etal-2019-stacl}, and (iii) the reinforcement learning (RL) policy~\cite{gu-etal-2017-learning}. The first approach offers a heuristically guided latency while the second one fixes it to an arbitrary value. The third one learns a stochastic policy to find the desired quality-latency trade-off. But before going into full details of methods, we now introduce the common metrics used to measure the latency of a given SiMT model.

\subsubsection{Latency metrics}
\label{sec:method_latency}
\B{Average proportion} (AP) is the very first metric used for latency measurement in the literature~\cite{cho2016can}. AP computes a normalised score between $0$ and $1$, which is the average number of source tokens required to commit a translation:
\begin{align*}
\text{AP}(X, Y) = \frac{1}{|X||Y|} \sum_{t=1}^{|Y|} g(t) 
\end{align*}
AP produces different scores for 2 samples when the underlying latency is actually the same but the source and target sentence lengths differ. To remedy this, \citet{ma-etal-2019-stacl} propose \B{Average Lagging} (AL) which estimates the number of tokens the ``writer'' is lagging behind the ``reader'', as a function of the number of input tokens read. $\tau$ denotes the timestep where the entire source sentence has been read, as the authors state that the subsequent timesteps do not incur further delay:
\begin{align*}
\text{AL}(X, Y) &= \frac{1}{\tau} \sum_{t=1}^{\tau} g(t) - \frac{t - 1}{\gamma}\quad (\gamma = \tfrac{|Y|}{|X|})
\end{align*}
Finally, \B{Consecutive Wait} (CW)~\cite{gu-etal-2017-learning} measures how many source tokens are consecutively read between committing two translations:
\begin{align*}
    C_0 &= 0 \\
    C_t &= 
    \begin{cases}
    C_{t-1} + 1 & \text{if action is \T{READ}}\\
    0           & \text{if action is \T{WRITE}}
    \end{cases}
\end{align*}

\subsubsection{Wait-$k$ training}
\label{sec:method_waitk}
\citet{ma-etal-2019-stacl} propose a simple deterministic policy which relies on training and decoding an NMT model in a \I{prefix-to-prefix} fashion.
Specifically, a wait-$k$ model starts by reading $k$ source tokens and writes the first target token. The model then reads and writes one token at a time to complete the translation process. This implies that the attention layer will now attend to a \B{partial textual representation} $H_{\le g(t)}$ instead of $H$,  with $g(t)$ redefined as $\min(k + t - 1, |X|)$ (eq.~\ref{eq:ctx} and ~\ref{eq:mle}).

\paragraph{Decoding-only mode.} A wait-$k$ model is denoted as ``trained'' if it is both trained and decoded using the algorithm above. It is also possible to take a pre-trained consecutive NMT or MMT model, and apply wait-$k$ algorithm at decoding time \ie during \I{greedy search}.

\subsubsection{Wait-if decoding}
\label{sec:method_waitif}
\citet{cho2016can} propose two decoding algorithms which can be directly applied on a pre-trained \I{consecutive} NMT model, similar to the consecutive wait-$k$ decoding. These algorithms have two hyper-parameters, namely the number of initial source tokens to read ($k$) before starting the translation and the number of further tokens to read ($\delta$) if the algorithm decides to wait for more context. We specifically use the \B{wait-if-diff (WID)} variant, which reads more tokens if the current most likely target word \B{changes} when doing so. We intentionally left out the \B{wait-if-worse (WIW)} approach as it exhibits very high latency.

\subsubsection{Reinforcement learning}
\label{sec:method_rl}
\citet{gu-etal-2017-learning} frame SiMT as a sequence of \T{READ} or \T{WRITE} actions and aim to learn a reinforcement learning (RL) strategy with a reward function taking into account both quality and latency. Following standard RL, the framework is composed of an environment and an agent. The environment is a pre-trained NMT system which is not \I{updated} during RL training. The agent is a GRU that parameterises a stochastic policy which decides on the action $a_t$ by receiving as input the observation $o_t$. The observation $o_t$ is defined as $\left[c_t; d_t; y_t\right]$, \ie the concatenation of vectors coming from the environment. At each step, the agent receives a reward $r_t = r_t^Q + r_t^D$ where $r_t^Q$ is the quality reward (the difference of smoothed BLEU scores for partial hypotheses produced from one step to another) and $r_t^D$ is the latency reward formulated as:
\begin{align*}
    \label{eq_rd}
    r_t^{D} = \alpha \left[\text{sgn}(C_t - C^*)+1\right] + \beta \lfloor D_t - D^* \rfloor_{+}
    \vspace{-3pt}
\end{align*}
\noindent where $C_t$ denotes the CW metric introduced here to avoid long consecutive waits and $D_t$ refers to AP (see $\S$~\ref{sec:method_latency} for metrics). $D^*$ and $C^*$ are hyper-parameters that determine the expected/target values for AP and CW, respectively. The optimal quality-latency trade-off is achieved by balancing the two reward terms.

\section{Experimental Setup}
\label{sec:setup}
\subsection{Dataset}
\begin{table}[t!]
\centering
\resizebox{.4\textwidth}{!}{
\begin{tabular}{@{}rlcccc@{}}
\toprule
           && \MC{2}{c}{En\ra De} & \MC{2}{c}{En\ra Fr} \\
           & Sents &  \MC{1}{c}{T/S}      & \MC{1}{c}{OOV}      & \MC{1}{c}{T/S}      & \MC{1}{c}{OOV}      \\
\midrule
\small TRAIN & 29K  & 0.96 & --     &1.10 & --  \\
\small VAL   & 1014 & 0.98 & 30.4\% &1.09 & 19.0\% \\
\small 2016  & 1000 & 0.94 &28.6\%  &1.09 & 18.1\% \\
\small 2017  & 1000 & 0.97 &37.1\%  &1.14 & 21.3\% \\
\small COCO  & 461  & 0.99 &38.2\%  &1.10 & 21.1\% \\ \bottomrule
\end{tabular}
}
\caption{Multi30k statistics: T/S and OOV are the average target-to-source sentence length ratio and the \% of sentence pairs with at least 1 unknown token.}
\label{tbl:data_stats}
\end{table}

We use the Multi30k dataset~\cite{elliott-etal-2016-multi30k}\footnote{https://github.com/multi30k/dataset} which has been the primary corpus for MMT research across the three shared tasks of the ``Conference on Machine Translation (WMT)'' ~\cite{specia-etal-2016-shared,elliott-etal-2017-findings,barrault-etal-2018-findings}. Multi30k extends the Flickr30k image captioning dataset~\cite{young-etal-2014-image} to provide caption translations in German, French and Czech. In this work, we focus on the English\ra German and English\ra French~\cite{elliott-etal-2017-findings} language directions (Table~\ref{tbl:data_stats}). We use flickr2016 (\B{2016}), flickr2017 (\B{2017}) and coco2017 (\B{COCO}) for model evaluation. The latter test set is explicitly designed~\cite{elliott-etal-2017-findings} to contain at least one ambiguous word per sentence, which makes it appealing for MMT experiments.

\paragraph{Preprocessing.} We use Moses scripts~\cite{koehn-etal-2007-moses} to lowercase, punctuation-normalise and tokenise the sentences with \I{hyphen splitting}. We then create word vocabularies on the \I{training} subset of the dataset. We did not use subword segmentation to avoid its potential side effects on SiMT and to be able to analyse the grounding capability of the models better. The resulting English, French and German vocabularies contain 9.8K, 11K and 18K tokens, respectively.

\subsection{Reproducibility}
\paragraph{Hyperparameters.}
The dimensions of embeddings and GRU hidden states are set to 200 and 320, respectively. The decoder's input and output embeddings are shared~\cite{press-wolf-2017-using}. We use ADAM~\cite{kingma2014adam} as the optimiser and set the learning rate and mini-batch size to 0.0004 and 64, respectively. A weight decay of $1e\rm{-}5$ is applied for regularisation. We clip the gradients if the norm of the full parameter vector exceeds $1$~\cite{pascanu2013difficulty}.
For the \B{RL baseline}, we closely follow~\cite{gu-etal-2017-learning}\footnote{https://github.com/nyu-dl/dl4mt-simul-trans}. The agent is implemented by a 320-dimensional GRU followed by a softmax layer and the baseline network -- used for variance reduction of policy gradient -- is similar to the agent except with a scalar output layer. We use ADAM as the optimiser and set the learning rate and mini-batch size to 0.0004 and 6, respectively. For each sentence pair in a batch, ten trajectories are sampled. For inference, greedy sampling is used to pick action sequences. We set the hyper-parameters $C^*\mathrm{=}2$, $D^*\mathrm{=}0.3$, $\alpha\mathrm{=}0.025$ and $\beta\mathrm{=}-1$. To encourage exploration, the negative entropy policy term is weighed empirically with 0.1 and 0.3 for En\ra Fr and En\ra De directions, respectively.

\paragraph{Training.} We use \T{nmtpytorch}~\cite{nmtpy} with PyTorch~\cite{paszke2017automatic} v1.4 for our experiments\footnote{https://github.com/ImperialNLP/pysimt}. We train each model for a maximum of 50 epochs and early stop the training if validation BLEU~\cite{papineni-etal-2002-bleu} does not improve for 10 epochs. We also halve the learning rate if no improvement is obtained for two epochs. On a single NVIDIA RTX2080-Ti GPU, it takes around 35 minutes for the \I{unimodal} and \I{multimodal encoder} variants to complete training whereas the \I{decoder attention} variant requires around twice that time. The number of learnable parameters is between 6.9M and 9.4M depending on the language pair and the type of multimodality.
For the \B{RL baseline}, we choose the model that maximises the quality-to-latency ratio (BLEU/AL) on the validation set with patience set to ten epochs. The number of learnable parameters is around 6M.

\subsection{Evaluation}
To mitigate variance in results due to different initialisations, we repeat each experiment \B{three} times, with random seeds. Following previous work, we decode translations with \I{greedy search}, using the checkpoint that achieved the lowest perplexity. We report \B{average} BLEU scores across three runs using sacreBLEU~\cite{post-2018-call}, which is also used for computing sentence-level scores for the oracle experiments.

\section{Results}
\label{sec:results}
\begin{table}[t]
\centering
\resizebox{.5\textwidth}{!}{
\begin{tabular}{@{}lccccccccc@{}}
\toprule
        & \MC{3}{c}{English\ra German} & \MC{3}{c}{English\ra French} & \\
        & 2016 & 2017 & \small COCO & 2016 & 2017 & \small COCO & Avg \\
\midrule
\small NMT      & 34.6  & 26.4   & 22.1  &
           57.8  & 50.3   & 41.4  &  --     \\
\midrule
\small ENC-OD   & \da{0.6} & \ua{0.3} & \ua{0.8} &  
           \ua{0.3} & \ua{0.1} & \ua{1.1} &  \ua{0.33}\\
\small DEC-OC   & \ua{0.4} & \ua{0.8} & \ua{0.7} &  
           \da{0.3} & \ua{0.2} & \ua{0.7} &  \ua{0.52}\\
\small DEC-OD   & \ua{0.7} & \ua{1.0} & \ua{1.7} & 
           \ua{0.1} & \ua{0.6} & \ua{1.2} &  \ua{0.88}\\
\bottomrule
\end{tabular}
}
\caption{Multimodal gains in BLEU for \I{consecutive} baselines: the DEC-OD system exhibits the best average improvements.}
\label{tbl:consecutive_bleu}
\end{table}

\subsection{Consecutive baselines}
We first present the impact of the visual integration approaches on \I{consecutive} NMT performance (Table~\ref{tbl:consecutive_bleu}). We observe that the decoder-attention using object detection features (DEC-OD) performs better than other variants. We also see that the improvements on flickr2017 (\ua{0.5}) and coco2017 (\ua{1.03}) test sets are higher than flickr2016 (\ua{0.1}) on average. A possible explanation is that flickr2017 and coco2017 are more distant from the training set distribution (higher OOV count, see Table~\ref{tbl:data_stats}) and thus there is more room for improvement with the visual cues. In summary, unlike previous conclusions in MMT where improvements were not found to be substantial~\cite{gronroos-etal-2018-memad, caglayan-etal-2019-probing}, we observe that the benefit of the visual modality is more pronounced here. We believe that this is due to (i) the encoder being now \I{unidirectional} different from state-of-the-art NMT, (ii) the modality representations being passed through layer normalisation~\cite{ba2016layer}, and (iii) the representational power of OD features.

\subsection{Unimodal SiMT baselines}

\begin{figure}[t]
\centering
\includegraphics[width=0.49\textwidth]{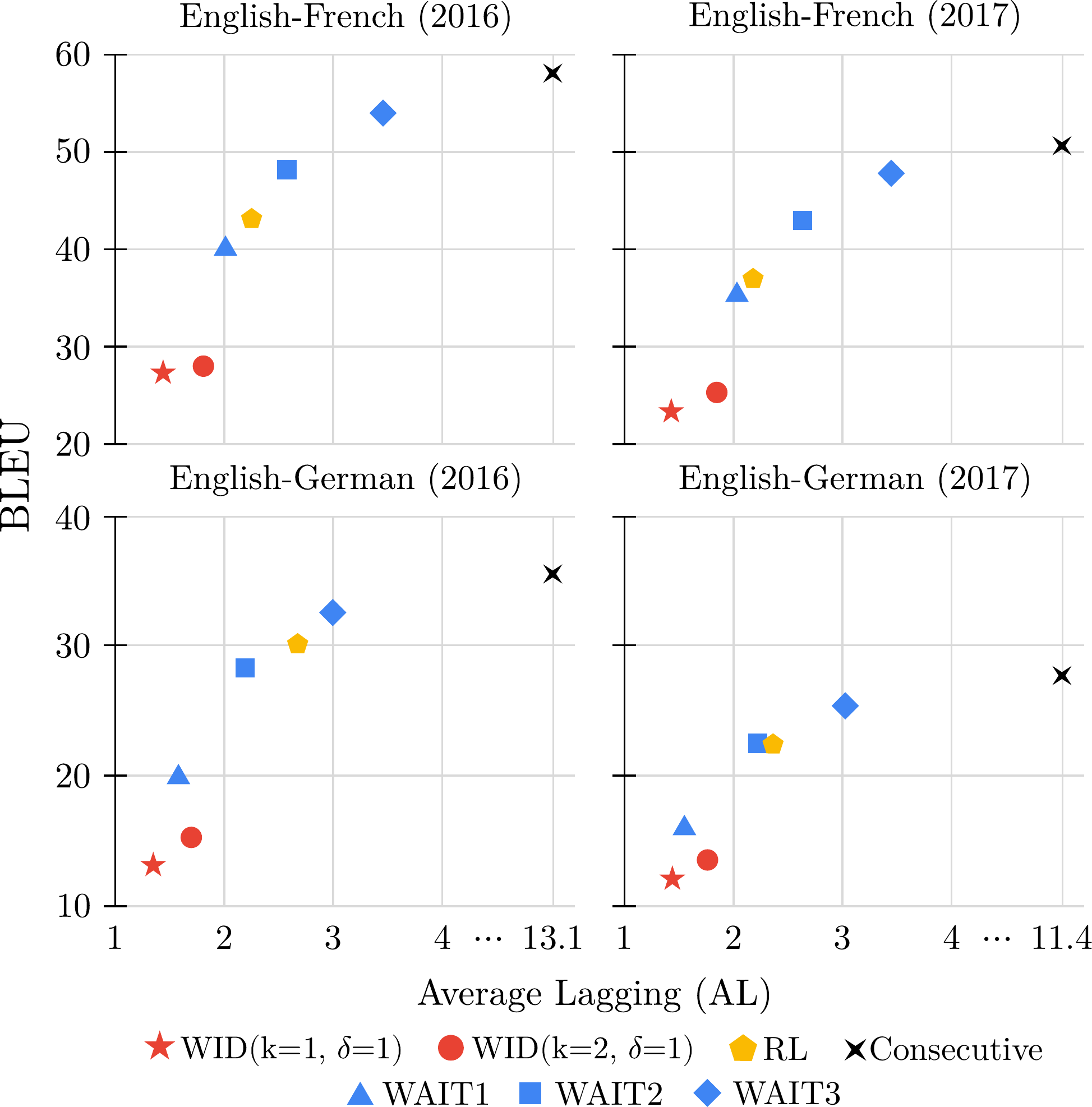}
\caption{AL vs BLEU comparison across unimodal SiMT approaches: wait-$k$ systems are ``trained''.}
\label{fig:rl_comparison}
\end{figure}

We now compare \B{unimodal} SiMT approaches to get an initial understanding of how they perform on Multi30k. Figure~\ref{fig:rl_comparison} contrasts AL and BLEU for three trained wait-$k$ systems, wait-if-diff (WID) decoding with $k \in \{1,2\}$ and $\delta\mathrm{=}1$, reinforcement learning (RL) and the consecutive NMT. These configurations are chosen particularly to satisfy a low-latency regimen. The results suggest that
wait-$k$ models offer good translation quality for fixed latency. The RL based policy, however, is not able to surpass wait-$k$ models. Finally, WID decoding exhibits the worst performance, according to BLEU. Given the difficulty in finding stable hyper-parameters for the RL models, we leave the integration of RL to MMT for future work and explore wait-$k$, and WID approaches in what follows.

\subsection{Wait-$k$ training for MMT}
\label{sec:results_trained_waitk}
\begin{figure}[t]
\centering
\includegraphics[width=0.48\textwidth]{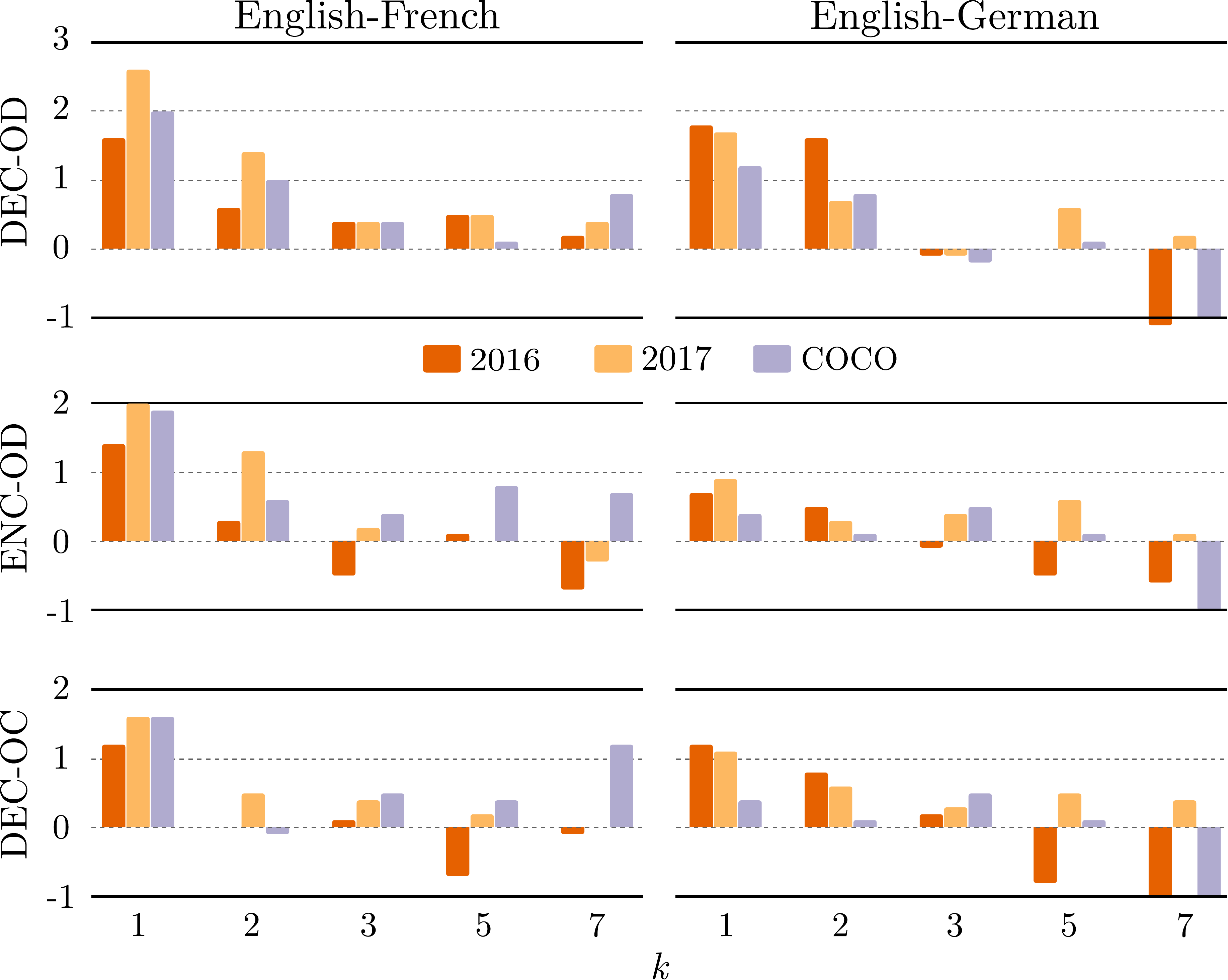}
\caption{BLEU comparison of trained wait-$k$ MMT systems: the vertical axes of each subplot represent the improvement with respect to the unimodal wait-$k$.}
\label{fig:trained_waitk_all}
\end{figure}

We present results with trained wait-$k$ MMTs with $k \in \{1,2,3,5,7\}$. Figure~\ref{fig:trained_waitk_all} plots a summary of the gains obtained by three MMT variants with respect to the unimodal wait-$k$. We observe that as $k$ increases, the gains due to the visual modality decrease globally, in line with the findings of \citet{caglayan-etal-2019-probing}. This phenomenon is more visible for German, which
exhibits 1 BLEU point drop consistently across all models and two test sets at $k\rm{=}7$. We hypothesise that this instability is probably due to the interplay of several factors for German, including the high OOV rates \& rich morphology and source sentences being slightly longer than target on average. The latter is a major issue for trained wait-$k$ systems since
the source sentences may not have been fully observed during training,\footnote{\citet{ma-etal-2019-stacl} proposed optional ``catchup'' logic for this, but we did not apply it here for the sake of simplicity.} preventing the decoder to learn about source \T{<EOS>} markers.
For French, the results are much more encouraging as the improvements are larger and still observed with larger $k$ values.

From a multimodal perspective, like with the consecutive models, the DEC-OD system has the best performance: it is beneficial for all values of $k$ in French, and it shows the largest gains in German for $k \in \{1,2\}$. From a runtime perspective, encoder-based attention benefits heavily from batched matrix operations and runs almost at the same speed as a unimodal NMT, thus encouraging us to focus more on that in the future.

\begin{table}[t!]
\renewcommand{\arraystretch}{1.0}
\centering
\resizebox{.4\textwidth}{!}{%
\begin{tabular}{@{}lr}
\toprule
& \MR{5}{*}{\includegraphics[height=2cm]{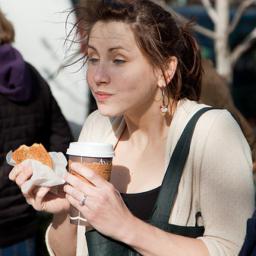}} \\
\T{SRC:} a young brunette woman ... & \\
\T{NMT:} \false{ein junger} \B{brünette frau ...} & \\
\T{MMT:} \true{eine junge} \B{brünette frau ...} & \\
\\
\midrule
& \MR{5}{*}{\includegraphics[height=2cm]{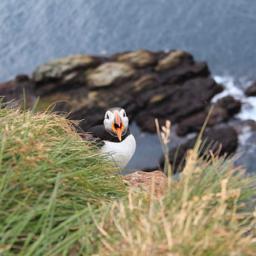}} \\
\T{SRC:} a black and white bird ... & \\
\T{NMT:} \B{un} \false{chien (dog)} \B{noir et blanc ...} & \\
\T{MMT:} \B{un} \true{oiseau (bird)} \B{noir et blanc ...} & \\
\\
\bottomrule
\end{tabular}}
\caption{Examples showing the effectiveness of DEC-OD MMT (wait-1) for gender marking (top) and adjective noun placement (bottom). \U{underlined} and \B{bold} represent wrong and correct word choices, respectively.}
\label{tbl:img_examples}
\end{table}
\paragraph{Qualitative examples.}
\label{sec:results_qual}
Table~\ref{tbl:img_examples} shows some examples regarding the impact of the visual modality for the wait-$1$ policy. In the first example, the image assists in predicting the correct article \B{eine} (feminine `a') instead of \B{ein} (masculine `a') in German. Upon inspecting the attention over object regions, we observe that the region that obtained the highest probability ($p$=0.2) when predicting \B{eine} is labelled with `woman' by the object detector. In the second example, we observe a biased anticipation case where the NMT system had to emit a wrong translation \B{chien} (`dog') before seeing the noun `bird'. However, the multimodal model successfully leveraged the visual context for anticipation and correctly handled the adjective-noun placement phenomenon. Once again, the attention distribution confirms that when generating the first two words -- \B{un} and \B{oiseau} (`bird'), the model correctly attends to the object regions corresponding to `bird' (with $p$=0.22 and $p$=0.14 respectively).

\subsection{Trained vs.\ decoding-only SiMT}
\begin{figure*}[t!]
\centering
\includegraphics[width=0.7\textwidth]{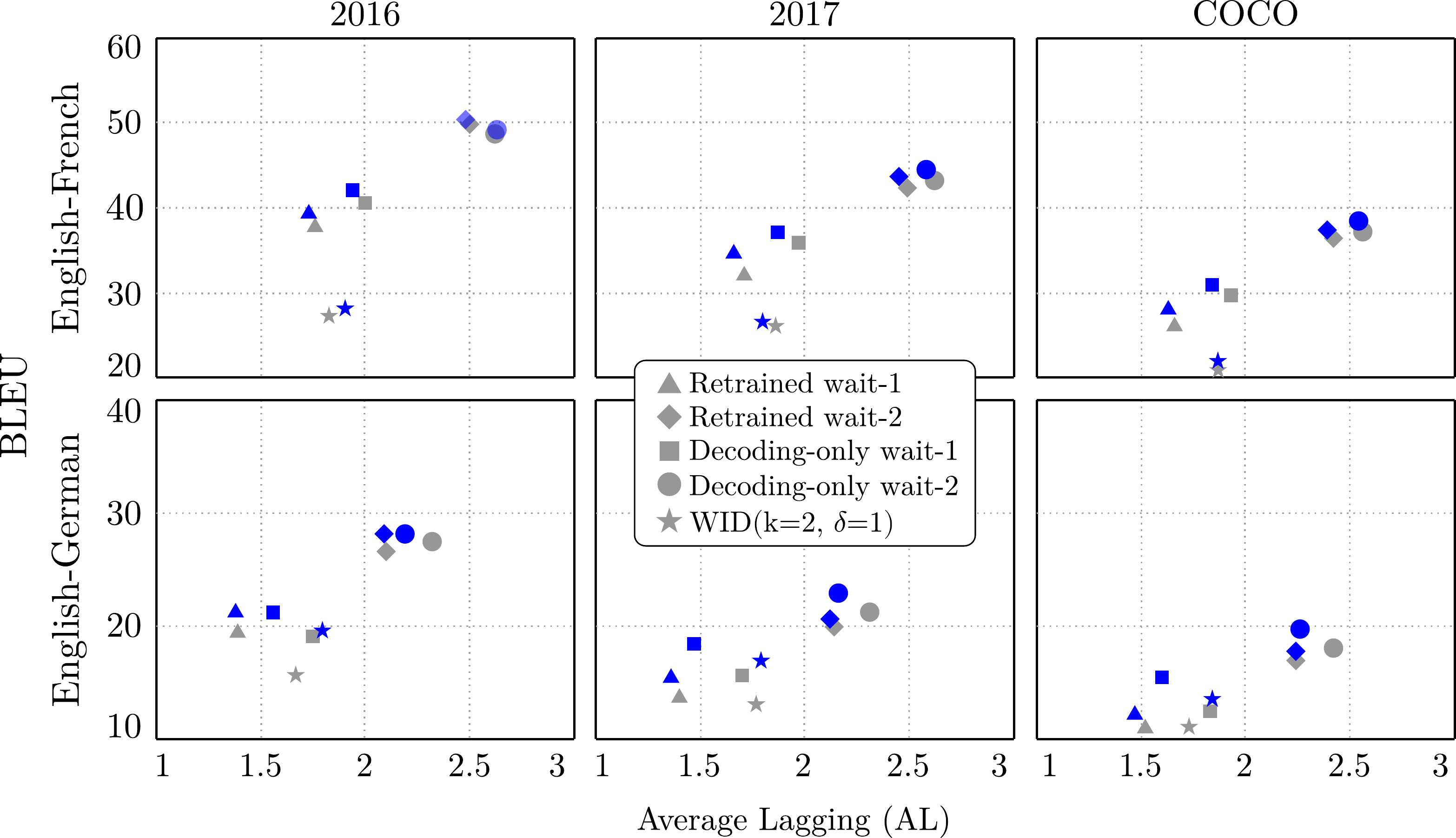}
\caption{Comparison of trained vs.\ decoding-only SiMT approaches: \textcolor{gray}{light} and \textcolor{blue}{dark} colors denote unimodal and multimodal (DEC-OD) systems, respectively.}
\label{fig:results_delay}
\end{figure*}

We are now interested in how trained wait-$k$ MMTs compare to decoding-only wait-$k$ and wait-if-diff (WID) heuristic under low latency. Figure~\ref{fig:results_delay} summarises latency vs.\ quality trade-off across all languages and test sets. First of all, the translation \B{quality} of the heuristic WID approach consistently improves with visual information with its latency slightly increasing across the board. Second, the translation \B{quality} of both trained and decoding-only wait-$k$ policies improve with multimodality.

Interestingly, although \citet{ma-etal-2019-stacl} show that trained wait-$k$ models are substantially better than decoding-only ones for a news translation task, here we observe quite the opposite: in almost all cases there exists a shift between these approaches which favours the decoding-only approach for small $k$ values. \citet{zheng-etal-2019-simpler} observed a similar phenomenon for their wait-1 and wait-2 textual SiMT models.
To investigate further, we compute an adaptive low-latency wait-$k$ \B{oracle} for $k \in \{1,2,3\}$. Specifically, for a given model, we first select a representative hypothesis across the three runs using median\footnote{We use median across the three runs of the same model as a way to smooth out variance related to random initialisations.} sentence-level BLEU. We then pick the hypothesis with best BLEU (and lowest $k$ in case of a tie) across all wait-$k$ systems of that model, as the oracle translation. Once the hypotheses are collected, we compute corpus-level BLEU and AL (Algorithm~\ref{alg:oracle}).

Table~\ref{tbl:oracles} suggests that when we let the oracle choose between different $k$ values, trained wait-$k$ systems are almost always better than decoding-only counterparts. Moreover, this boost in quality is accompanied by slight latency improvements over unimodal NMT across the board. Therefore, we conjecture that the shifts between decoding-only and trained wait-$k$ systems may be due to several factors coming into play such as the length discrepancy issue ($\S$\ref{sec:results_trained_waitk}) or the low resourced nature of Multi30k which prevents it from benefiting from prefix-to-prefix training for small $k$ values.

\begin{algorithm}[t]
\small
\SetInd{0.5em}{0.5em}
\SetKwInOut{Input}{inputs}
\SetKwInOut{Output}{output}
\Output{Oracle BLEU $\rightarrow$ CorpusBLEU$(O)$}
\Output{Oracle Delay $\rightarrow$ AverageLag$(O)$}

$N :$ Number of test set sentences\\
$B :$ Sentence BLEU scores across runs\\
$C :$ Candidate list of \{HYP, BLEU, $k$\}\\
$O :$ Final oracle list of $N$ translations\\

$O \leftarrow$ []\\
\For{n in $1\dots N$}{
    $C \leftarrow$ []\\
    \For{k in $\{1,2,3,5,7\}$}{
        $R\leftarrow$ wait-$k$ hypotheses for input $n$\\
        $B\leftarrow [\text{BLEU}\left(R[i]\right) \forall i \in \{1,2,3\}]$\\
        $m\leftarrow $ Index of run with median BLEU\\
        HYP $\leftarrow R[m]$;\, BLEU $\leftarrow B[m]$\\
        $C$.append(\{HYP, BLEU, k\})\\
    }
    $h\leftarrow$ Best HYP from $C$ (lowest $k$ if tie)\\
    $O$.append($h$)\\
}
\caption{Multi-run oracle algorithm}
\label{alg:oracle}
\end{algorithm}
\begin{table}[t!]
\centering
\resizebox{.42\textwidth}{!}{
\begin{tabular}{@{}lccc@{}}
\toprule
    & 2016  & 2017 & COCO \\ \cmidrule(l){2-4}
    & \MC{3}{c}{English\ra German} \\
\midrule
\MR{2}{*}{NMT}  & 33.5 (2.52)    & 26.2 (2.46)    & 21.9 (2.61) \\
                & 33.7 (2.30)    & 25.6 (2.32)    & 22.4 \B{(2.38)}    \\ 
\midrule
\MR{2}{*}{DEC-OD} & \B{34.4} (2.37)   & \B{27.8} (2.24)   & \B{23.8} (2.40) \\
                  & 34.3 \B{(2.23)}   & 26.0 \B{(2.20)}   & 22.5 (2.45)    \\
\midrule
    & \MC{3}{c}{English\ra French} \\
\midrule
\MR{2}{*}{NMT}  &55.3 (2.78)  &48.9 (2.71)  & 41.4 (2.62)    \\
                & 56.9 (2.69) & 50.4 (2.69)    & 42.4 (2.68)   \\ 
\midrule
\MR{2}{*}{DEC-OD}&55.9 (2.71)  &50.1 \B{(2.58)}  & 42.3 (2.58)    \\
                & \B{57.9 (2.65)}    & \B{51.3} (2.60)    & \B{43.1 (2.57)}   \\
\bottomrule
\end{tabular}
}
\caption{BLEU (AL) oracles for low-latency decoding-only (first line) and trained wait-$k$ (second line).}
\label{tbl:oracles}
\end{table}

\paragraph{Gender resolution accuracy.}
Motivated by the qualitative examples in $\S$\ref{sec:results_qual}, 
we further investigate how accurate English\ra French MMT variants are at choosing the correct indefinite article \B{une} when translating sentences beginning with `a woman'. Table~\ref{tbl:femme_stats} shows that the unimodal NMT has no way of anticipating the context to resolve this kind of gender ambiguity, and therefore always picks the masculine version of the article. This is a clear example of models reflecting biases in the training data. In fact, 69.4\% of all training instances starting with an indefinite article in French, have the masculine realisation of the article (\B{un}) instead of its feminine counterpart (\B{une}). The results also make it clear that decoding-only wait-$k$ systems are not as successful as their trained counterparts when it comes to incorporating the visual modality, and the explicit object information is more expressive than global object features. At $k=2$ however, all systems reach 100\% accuracy eventually.

\begin{table}[t]
\centering
\renewcommand{\arraystretch}{1.0}
\resizebox{.35\textwidth}{!}{
\begin{tabular}{@{}rcc@{}}
\toprule
wait-1 & \small DECODING ONLY & \small TRAINED \\
\midrule
\small NMT     & 0              & 0         \\
\small DEC-OC  & 10.7           & 49        \\
\small ENC-OD  & \phantom{1}7.9 & 71        \\
\small DEC-OD  & \B{18.7}       & \B{72}    \\
\bottomrule
\end{tabular}}
\caption{Gender resolution accuracy of decoding only and retrained wait-1 systems when translating sentences starting with `a woman ...' into French.}
\label{tbl:femme_stats}
\end{table}

\section{Conclusion}
\label{sec:conclusion}
We present the first thorough investigation of the utility of visual context for the task of simultaneous machine translation. Our experiments reveal that integrating visual context lowers the latency for heuristic policies while retaining the quality of the translations. Under low-latency wait-$k$ policies, the visual cues are highly impactful with quality improvements of almost $3$ BLEU points compared to unimodal baselines. From a multimodal perspective, we introduce effective ways of integrating visual features and show that explicit object region information consistently outperforms commonly used global features. Our qualitative analysis illustrates that the models are capable of resolving linguistic particularities, including gender marking and word order handling by exploiting the associated visual cues.

We hope that future research continues this line of work, especially by finding novel ways to devise adaptive policies -- such as reinforcement learning models with the visual modality. We believe that our work can also benefit research in multimodal speech translation~\cite{iwslt19} where the audio stream is accompanied by a video stream.

\section*{Acknowledgments}
We would like to thank Aizhan Imankulova for being very responsive to our questions regarding their results in~\citet{imankulova2020multimodal}, and Kyunghyun Cho for answering a question related to wait-if decoding~\cite{cho2016can}.

Ozan Caglayan, Julia Ive, Pranava Madhyastha and Lucia Specia received funding from the MultiMT (H2020 ERC Starting Grant No. 678017) and MMVC (Newton Fund Institutional Links Grant, ID 352343575) projects.

\bibliography{anthology,paper}
\bibliographystyle{acl_natbib}

\end{document}